\documentclass[submission,copyright,creativecommons,sharealike,noncommercial]{eptcs}
\providecommand{\event}{SemSpace 2020}
\pdfoutput=1
\usepackage{amsmath}
\usepackage{amsthm}
\usepackage{amssymb}
\usepackage{graphicx}
\usepackage{wrapfig}
\usepackage{url}
\usepackage{mathtools, nccmath}
\usepackage[font={sf}]{caption}
\usepackage{tikz}
\usepackage{tikzit}
\usepackage{fontawesome}
\usepackage{stmaryrd}

\tikzstyle{doubled}=[line width=1.5pt] % set the line width for all doubled (quantum) maps/wires

\tikzstyle{dot}=[inner sep=0mm,minimum width=2mm,minimum height=2mm,draw,shape=circle]  
\tikzstyle{ddot}=[inner sep=0mm, doubled, minimum width=2.5mm,minimum height=2.5mm,draw,shape=circle]

\tikzstyle{pdot}=[inner sep=0mm, doubled, minimum width=2.5mm,minimum height=2.5mm,shape=circle]
\tikzstyle{phase dimensions}=[minimum size=6mm,font=\footnotesize,inner sep=0.2mm,outer sep=-2mm]

\tikzstyle{phase dot}=[pdot,phase dimensions]
\tikzstyle{wphase dot}=[dot, phase dimensions]

\tikzstyle{hadamard}=[fill=white,draw,inner sep=0.6mm,font=\footnotesize,minimum height=6mm,minimum width=8mm]

\tikzstyle{anti} = [fill=white,draw,inner sep=0.6mm,font=\footnotesize,minimum height=3mm,minimum width=3mm]

\tikzstyle{triang}=[regular polygon,regular polygon sides=3,draw,scale=0.75,inner sep=-0.75pt,minimum width=9mm,fill=white,regular polygon rotate=180]
\tikzstyle{triang_lesssep}=[regular polygon,regular polygon sides=3,draw,scale=0.75,inner sep=-4pt,minimum width=9mm,fill=white,regular polygon rotate=180, text depth=4mm]
\tikzstyle{triangdag}=[regular polygon,regular polygon sides=3,draw,scale=0.75,inner sep=-0.5pt,minimum width=9mm,fill=white]

\makeatletter
\newcommand{\boxshape}[3]{%
\pgfdeclareshape{#1}{
\inheritsavedanchors[from=rectangle] % this is nearly a rectangle
\inheritanchorborder[from=rectangle]
\inheritanchor[from=rectangle]{center}
\inheritanchor[from=rectangle]{north}
\inheritanchor[from=rectangle]{south}
\inheritanchor[from=rectangle]{west}
\inheritanchor[from=rectangle]{east}
\backgroundpath{% this is new
\southwest \pgf@xa=\pgf@x \pgf@ya=\pgf@y
\northeast \pgf@xb=\pgf@x \pgf@yb=\pgf@y

\@tempdima=#2
\@tempdimb=#3

\pgfpathmoveto{\pgfpoint{\pgf@xa - 5pt + \@tempdima}{\pgf@ya}}
\pgfpathlineto{\pgfpoint{\pgf@xa - 5pt - \@tempdima}{\pgf@yb}}
\pgfpathlineto{\pgfpoint{\pgf@xb + 5pt + \@tempdimb}{\pgf@yb}}
\pgfpathlineto{\pgfpoint{\pgf@xb + 5pt - \@tempdimb}{\pgf@ya}}
\pgfpathlineto{\pgfpoint{\pgf@xa - 5pt + \@tempdima}{\pgf@ya}}
\pgfpathclose
}
}}

\boxshape{NEbox}{0pt}{5pt}
\boxshape{SEbox}{0pt}{-5pt}
\boxshape{NWbox}{5pt}{0pt}
\boxshape{SWbox}{-5pt}{0pt}
\boxshape{EBox}{-3pt}{3pt}
\boxshape{WBox}{3pt}{-3pt}
\makeatother

\tikzstyle{map}=[draw,shape=NEbox,inner sep=2pt,minimum height=6mm,fill=white]
\tikzstyle{mapdag}=[draw,shape=SEbox,inner sep=2pt,minimum height=6mm,fill=white]
\tikzstyle{maptrans}=[draw,shape=SWbox,inner sep=2pt,minimum height=6mm,fill=white]
\tikzstyle{mapconj}=[draw,shape=NWbox,inner sep=2pt,minimum height=6mm,fill=white]

\tikzstyle{dmap}=[draw,doubled,shape=NEbox,inner sep=2pt,minimum height=6mm,fill=white]
\tikzstyle{dmapdag}=[draw,doubled,shape=SEbox,inner sep=2pt,minimum height=6mm,fill=white]
\tikzstyle{dmaptrans}=[draw,doubled,shape=SWbox,inner sep=2pt,minimum height=6mm,fill=white]
\tikzstyle{dmapconj}=[draw,doubled,shape=NWbox,inner sep=2pt,minimum height=6mm,fill=white]

\makeatletter
\pgfdeclareshape{cornerpoint}{
\inheritsavedanchors[from=rectangle] % this is nearly a rectangle
\inheritanchorborder[from=rectangle]
\inheritanchor[from=rectangle]{center}
\inheritanchor[from=rectangle]{north}
\inheritanchor[from=rectangle]{south}
\inheritanchor[from=rectangle]{west}
\inheritanchor[from=rectangle]{east}
\backgroundpath{% this is new
\southwest \pgf@xa=\pgf@x \pgf@ya=\pgf@y
\northeast \pgf@xb=\pgf@x \pgf@yb=\pgf@y

\pgfmathsetmacro{\pgf@shorten@left}{\pgfkeysvalueof{/tikz/shorten left}}
\pgfmathsetmacro{\pgf@shorten@right}{\pgfkeysvalueof{/tikz/shorten right}}

\pgfpathmoveto{\pgfpoint{0.5 * (\pgf@xa + \pgf@xb)}{\pgf@ya - 5pt}}
\pgfpathlineto{\pgfpoint{\pgf@xa - 8pt + \pgf@shorten@left}{\pgf@yb - 1.5 * \pgf@shorten@left}}
\pgfpathlineto{\pgfpoint{\pgf@xa - 8pt + \pgf@shorten@left}{\pgf@yb}}
\pgfpathlineto{\pgfpoint{\pgf@xb + 8pt - \pgf@shorten@right}{\pgf@yb}}
\pgfpathlineto{\pgfpoint{\pgf@xb + 8pt - \pgf@shorten@right}{\pgf@yb - 1.5 * \pgf@shorten@right}}
\pgfpathclose
}
}

\pgfdeclareshape{cornercopoint}{
\inheritsavedanchors[from=rectangle] % this is nearly a rectangle
\inheritanchorborder[from=rectangle]
\inheritanchor[from=rectangle]{center}
\inheritanchor[from=rectangle]{north}
\inheritanchor[from=rectangle]{south}
\inheritanchor[from=rectangle]{west}
\inheritanchor[from=rectangle]{east}
\backgroundpath{% this is new
\southwest \pgf@xa=\pgf@x \pgf@ya=\pgf@y
\northeast \pgf@xb=\pgf@x \pgf@yb=\pgf@y

\pgfmathsetmacro{\pgf@shorten@left}{\pgfkeysvalueof{/tikz/shorten left}}
\pgfmathsetmacro{\pgf@shorten@right}{\pgfkeysvalueof{/tikz/shorten right}}

\pgfpathmoveto{\pgfpoint{0.5 * (\pgf@xa + \pgf@xb)}{\pgf@yb + 5pt}}
\pgfpathlineto{\pgfpoint{\pgf@xa - 8pt + \pgf@shorten@left}{\pgf@ya + 1.5 * \pgf@shorten@left}}
\pgfpathlineto{\pgfpoint{\pgf@xa - 8pt + \pgf@shorten@left}{\pgf@ya}}
\pgfpathlineto{\pgfpoint{\pgf@xb + 8pt - \pgf@shorten@right}{\pgf@ya}}
\pgfpathlineto{\pgfpoint{\pgf@xb + 8pt - \pgf@shorten@right}{\pgf@ya + 1.5 * \pgf@shorten@right}}
\pgfpathclose
}
}

\makeatother

\pgfkeyssetvalue{/tikz/shorten left}{0pt}
\pgfkeyssetvalue{/tikz/shorten right}{0pt}

\tikzstyle{kpoint common}=[draw,fill=white,inner sep=1pt,minimum height=4mm]
\tikzstyle{kpoint}=[shape=cornerpoint,shorten left=5pt,kpoint common]
\tikzstyle{kpoint adjoint}=[shape=cornercopoint,shorten left=5pt,kpoint common]
\tikzstyle{kpoint conjugate}=[shape=cornerpoint,shorten right=5pt,kpoint common]
\tikzstyle{kpoint transpose}=[shape=cornercopoint,shorten right=5pt,kpoint common]

\tikzstyle{kpointdag}=[kpoint adjoint]
\tikzstyle{kpointadj}=[kpoint adjoint]
\tikzstyle{kpointconj}=[kpoint conjugate]
\tikzstyle{kpointtrans}=[kpoint transpose]

\tikzstyle{big kpoint}=[kpoint, minimum width=1.0 cm, minimum height=2mm, inner sep=4pt, text depth=1.5mm]

 \tikzstyle{upground}=[circuit ee IEC,thick,ground,rotate=90,scale=1.5]
 \tikzstyle{downground}=[circuit ee IEC,thick,ground,rotate=-90,scale=1.5]

\tikzstyle{discarding}=[fill=white, draw=black, shape=circle, style=upground]
\tikzstyle{smalldiscarding}=[fill=white, draw=black, style=upground, scale=0.5]
\tikzstyle{backdiscard}=[fill=white, draw=black, shape=circle, style=downground, scale=0.5]
\tikzstyle{smallbackdiscard}=[fill=white, draw=black, shape=circle, style=downground, scale=0.5]
\tikzstyle{state}=[fill=white, draw=black, style=triang, tikzit shape=rectangle]
\tikzstyle{kstate}=[fill=white, draw=black, style=kpoint, tikzit shape=rectangle]
\tikzstyle{kstateconj}=[fill=white, draw=black, style=kpoint conjugate, tikzit shape=rectangle]
\tikzstyle{kstateBIG}=[fill=white, draw=black, style=big kpoint, tikzit shape=rectangle]
\tikzstyle{effect}=[fill=white, draw=black, style=triangdag]
\tikzstyle{keffect}=[fill=white, draw=black, style=kpoint adjoint]
\tikzstyle{keffectconj}=[fill=white, draw=black, style=kpoint transpose]
\tikzstyle{morphdag}=[style=mapdag]
\tikzstyle{morph}=[style=hadamard]
\tikzstyle{WIDEmorph}=[style=hadamard, minimum width=14mm]
\tikzstyle{morphtrans}=[style=maptrans]
\tikzstyle{morphconj}=[style=mapconj]
\tikzstyle{CPMmorph}=[style=dmap]
\tikzstyle{CPMmorphconj}=[style=dmapconj]
\tikzstyle{CPMmorphdag}=[style=dmapdag]
\tikzstyle{CPMmorphtrans}=[style=dmaptrans]
\tikzstyle{CPMstate}=[fill=white, draw=black, style=triang, doubled]
\tikzstyle{CPMstateBIG}=[fill=white, draw=black, style={triang_lesssep}, doubled]
\tikzstyle{CPMkstate}=[fill=white, draw=black, style=kpoint, tikzit shape=rectangle, doubled]
\tikzstyle{CPMkstateconj}=[fill=white, draw=black, style=kpoint conjugate, tikzit shape=rectangle, doubled]
\tikzstyle{CPMkstateBIG}=[fill=white, draw=black, style=big kpoint, tikzit shape=rectangle, doubled]
\tikzstyle{CPMkeffect}=[fill=white, draw=black, style=kpoint adjoint, doubled]
\tikzstyle{CPMkeffectconj}=[fill=white, draw=black, style=kpoint transpose, doubled]
\tikzstyle{UHfB}=[fill=white, draw=black, style=triangdag, doubled, inner sep=-2pt]
\tikzstyle{leak}=[style=tinypoint, regular polygon rotate=-90]
\tikzstyle{leakfill}=[style=tinypoint, regular polygon rotate=-90, fill=black]
\tikzstyle{Z}=[style=dot, fill=green]
\tikzstyle{X}=[style=dot, fill=red]
\tikzstyle{black_dot}=[style=dot, fill=black]
\tikzstyle{white_dot}=[style=dot, fill=white]
\tikzstyle{qblack_dot}=[style=ddot, fill=black]
\tikzstyle{qwhite_dot}=[style=ddot, fill=white]
\tikzstyle{whitephase}=[style=wphase dot, fill=white]
\tikzstyle{qredphase}=[style=phase dot, fill=red]
\tikzstyle{qgreenphase}=[style=phase dot, fill=green]
\tikzstyle{had}=[style=hadamard, doubled]
\tikzstyle{box}=[style=hadamard]
\tikzstyle{classhad}=[style=hadamard]
\tikzstyle{antipode}=[style=anti]

\tikzstyle{dottededge}=[-, dotted]
\tikzstyle{double edge}=[-, style=doubled, draw=black, tikzit draw={rgb,255: red,191; green,0; blue,64}]
\tikzstyle{new edge style 0}=[<-]

\usetikzlibrary{calc, positioning, shapes.geometric}
\usetikzlibrary{
	arrows,
	shapes,
	decorations,
	intersections,
	backgrounds,
	positioning,
	circuits.ee.IEC
	}

\newcommand{\tikzfigscale}[2]{\scalebox{#1}{\input{#2.tikz}}}

\newcommand{\cat}{\mathbf}
\newcommand{\morph}[1]{\xrightarrow{#1}}
\newcommand{\id}[1]{\textrm{id}_{#1}}
\newcommand{\ob}[1]{\textrm{Ob}(#1)}
\newcommand{\cor}{\operatorname{\rotatebox[origin=c]{90}{\text{\faRandom}}}}
\newcommand{\putt}[1]{\mathbin{\uparrow_{#1}}}
\newcommand{\get}[1]{\operatorname{\text{\faEye}}_{#1}}
\newcommand{\mix}[1]{\sim_{#1}}
\newcommand{\copyy}[1]{
\mathbin{\begin{tikzpicture}
		\node [style={black_dot}, scale=0.6] (0) at (0, 0) {};
		\node [style=none] (1) at (0, -0.2) {};
		\node [style=none] (2) at (-0.2, 0.2) {};
		\node [style=none] (3) at (0.2, 0.2) {};
		\draw (1.center) to (0);
		\draw [in=-90, out=150] (0) to (2.center);
		\draw [in=270, out=30] (0) to (3.center);
\end{tikzpicture}}_{\hspace{-0.05cm}#1}}

\newcommand{\wmult}{
\mathbin{\begin{tikzpicture}[rotate=180]
		\node [style={white_dot}, scale=0.6] (0) at (0, 0) {};
		\node [style=none] (1) at (0, -0.2) {};
		\node [style=none] (2) at (-0.2, 0.2) {};
		\node [style=none] (3) at (0.2, 0.2) {};
		\draw (1.center) to (0);
		\draw [in=-90, out=150] (0) to (2.center);
		\draw [in=270, out=30] (0) to (3.center);
\end{tikzpicture}}}

\newcommand{\bmult}{
\mathbin{\begin{tikzpicture}[rotate=180]
		\node [style={black_dot}, scale=0.6] (0) at (0, 0) {};
		\node [style=none] (1) at (0, -0.2) {};
		\node [style=none] (2) at (-0.2, 0.2) {};
		\node [style=none] (3) at (0.2, 0.2) {};
		\draw (1.center) to (0);
		\draw [in=-90, out=150] (0) to (2.center);
		\draw [in=270, out=30] (0) to (3.center);
\end{tikzpicture}}}

\newcommand{\wcomult}{
\mathbin{\begin{tikzpicture}
		\node [style={white_dot}, scale=0.6] (0) at (0, 0) {};
		\node [style=none] (1) at (0, -0.2) {};
		\node [style=none] (2) at (-0.2, 0.2) {};
		\node [style=none] (3) at (0.2, 0.2) {};
		\draw (1.center) to (0);
		\draw [in=-90, out=150] (0) to (2.center);
		\draw [in=270, out=30] (0) to (3.center);
\end{tikzpicture}}}

\newcommand{\bunit}{\,
\mathbin{\begin{tikzpicture}
	\begin{pgfonlayer}{nodelayer}
		\node [style={black_dot}, scale=0.6] (0) at (0, -0.1) {};
		\node [style=none] (1) at (0, 0.2) {};
	\end{pgfonlayer}
	\begin{pgfonlayer}{edgelayer}
		\draw (1.center) to (0);
	\end{pgfonlayer}
\end{tikzpicture}} \, }

\newcommand{\wunit}{\,
\mathbin{\begin{tikzpicture}
	\begin{pgfonlayer}{nodelayer}
		\node [style={white_dot}, scale=0.6] (0) at (0, -0.1) {};
		\node [style=none] (1) at (0, 0.2) {};
	\end{pgfonlayer}
	\begin{pgfonlayer}{edgelayer}
		\draw (1.center) to (0);
	\end{pgfonlayer}
\end{tikzpicture}} \, }

\newcommand{\banana}{\raisebox{-0.2cm}{\includegraphics[scale=0.03]{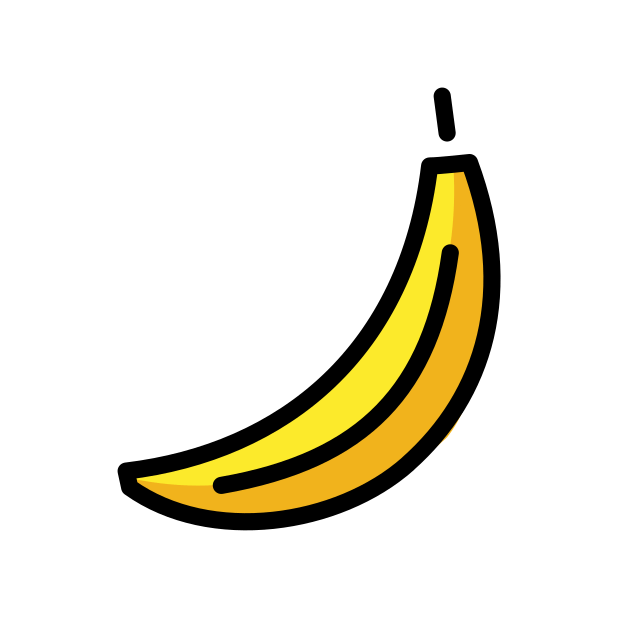}}}
\newcommand{\smallbanana}{\raisebox{-0.01cm}{\includegraphics[scale=0.013]{figs/banana}}}
\newcommand{\monkey}{\raisebox{-0.1cm}{\includegraphics[scale=0.05]{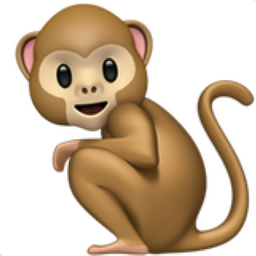}}}
\newcommand{\smallmonkey}{\raisebox{-0.01cm}{\includegraphics[scale=0.03]{figs/monkey}}}
\newcommand{\smiley}{\raisebox{-0.15cm}{\includegraphics[scale=0.03]{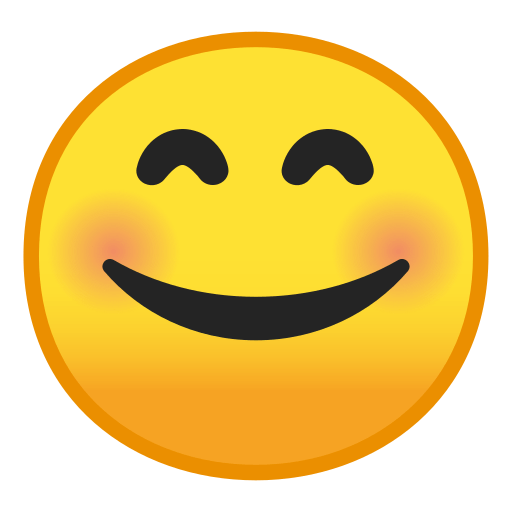}}}
\newcommand{\smallsmiley}{\raisebox{-0.01cm}{\includegraphics[scale=0.013]{figs/smile}}}

\theoremstyle{definition}
\newtheorem{defn}{Definition}
\theoremstyle{plain}
\newtheorem{prop}{Proposition}

\begin{document}

\title{Categories of Semantic Concepts}
\def\titlerunning{Categories of Semantic Concepts}
%\titlethanks{With thanks to Sean Tull, Cole Comfort, Irene Rizzo, and Bob Coecke for helpful discussions.
%This work was supported by University College London and the EPSRC
%[grant number EP/L015242/1].
%The authors contributed equally to this work.}
\author{
James Hefford \institute{University of Oxford} \email{james.hefford@cs.ox.ac.uk} \and 
Matthew Wilson \institute{University of Oxford} \institute{HKU-Oxford Joint Laboratory for \\ Quantum Information and Computation \email{matthew.wilson@cs.ox.ac.uk}} \and
Vincent Wang  \institute{University of Oxford} \email{vincent.wang@cs.ox.ac.uk}}
\def\authorrunning{J. Hefford, V. Wang and M. Wilson}

\maketitle

\begin{abstract}
    Modelling concept representation is a foundational problem in the study of cognition and linguistics. This work builds on the confluence of conceptual tools from G\"{a}rdenfors semantic spaces, categorical compositional linguistics, and applied category theory to present a domain-independent and categorical formalism of `concept'.
\end{abstract}

\section{Introduction}

G\"{a}rdenfors proposes geometry as a mediator between raw empirical data and symbolic cognitive representations of concepts \cite{gardenfors_conceptual_2004}. Conceptual spaces conceived in this way were instrumental in developing state-of-the-art models for semantic inference from linguistic data \cite{bolt_interacting_2017}, in conjunction with tools from Firth's distributional models of meaning, and the DisCoCat programme of categorical grammar and semantics \cite{coecke_mathematical_2010}.

G\"{a}rdenfors' model treats concepts as regions of abstract geometric domains that encode properties; for instance, where the domains are colour and taste, the concept `Banana' might be expressed as a well-behaved restricted region that captures the expected inferences that no bananas are red and sour, while yellow bananas tend to be sweet, and green ones bitter.

We develop towards a categorical formalism that builds categories of semantic concepts -- such as `Banana' -- on top of arbitrary categories, the objects of which are the modeller's choice of conceptual spaces, that constitute what G\"{a}rdenfors calls \emph{integral domains}.  We briefly recount G\"{a}rdenfors' ontology. An integral domain is a semantic space where one cannot have incomplete information, just as one cannot perceive the pitch of a sound without also perceiving the volume. Hue (a part of colourspace, along with brightness) is such an integral domain, representable with four quality dimensions: redness, greenness, blueness, and yellowness, subject to geometric constraints. Similarly, tastespace is an integral domain, along, say, the dimensions of sweetness, savouriness, bitterness, sourness, and umami. However, when considering fruitspace, modelled as the (tensored) conjunction of colour and taste, it is possible to specify colour without taste (by just looking), and taste without colour (by eating blindfolded). In this case we refer to the domains of colour and taste as independent domains. We abstractly deem a concept to be a subregion of possibly multiple tensorially conjoined independent domains. For instance, where fruitspace is the tensor of colourspace and tastespace, the concept `banana' falls within an acceptable range of yellowness and greenness, sweetness and bitterness, in a potentially non-trivial way: we may tolerate different taste-ranges in bananas depending on their colour. Percepts that stray too far outside these constraints may not be intuitively classified as bananas \emph{per se}, as sour purple bananas are like black swans: arguably not bananas at all.

In the remainder of this paper, we first briefly introduce symmetric monoidal categories and their attendant graphical calculus, the tool of choice for the DisCoCat school of modelling categorical semantics. We then develop a cognitively-driven axiomatic account of updating static properties of systems. We then introduce the notion of correlators on update structures, and by demonstration, we exhibit how concepts may be constructed from domains by their restriction and correlating actions. We then recover from these tools the first categorical formulation of G\"{a}rdenfors' semantic concepts, and show how our construction reconciles with our account of updating.

\section{Mathematical Background}

For a thorough account of basic category theory, we refer the reader to \cite{abramsky_introduction_2010}. For an account of the available graphical calculi for various monoidal categories, we refer to \cite{selinger_survey_2010}.

\begin{defn}[Category]
  A category $\cat{C}$ consists of the following data:
  \begin{itemize}
    \item{A collection $\ob{\cat{C}}$, the elements of which we call \textbf{objects} of $\cat{C}$.} 
    \item{For every pair of objects $A,B$, a collection $\cat{C}(A,B)$ of \textbf{morphisms}, or \textbf{arrows}, from $A$ to $B$. We might denote such a morphism $f: A \rightarrow B$, or $A \overset{f}{\rightarrow} B$.}
    \item{For every object $A$, a morphism $\id{A}:A\morph{}A$, called the \textbf{identity morphism} on $A$.}
    \item{For every triple of objects $A,B,C$, and every pair of morphisms $f:A\morph{}B$ and $g:B\morph{}C$, a specification of a morphism $g\circ f:A\morph{}C$ called the \textbf{composite} of $f$ and $g$. This composite is sometimes also written $f ; g$.}
  \end{itemize}
  Subject to the following conditions:
  \begin{itemize}
    \item{\emph{(Unitality)}: For any $A\overset{f}{\rightarrow}B$, we have $\id{A} ; f = f = f; \id{B}$; that is, composing with identities does nothing.}
    \item{\emph{(Associativity)}: For any three morphisms $A \overset{f}{\rightarrow}B$, $B \overset{g}{\rightarrow} C$, $C \overset{h}{\rightarrow}D$, we have $(f;g);h = f;(g;h)$.}
  \end{itemize}
\end{defn}

\begin{defn}[Monoidal Category]
  A monoidal category is a tuple $(\cat{C},\otimes,I,\alpha,\lambda,\rho)$ where:
  \begin{itemize}
    \item{$\cat{C}$ is a category}
    \item{$\otimes$, the \textbf{tensor}, is a functor $\cat{C} \times \cat{C} \rightarrow \cat{C}$. We use the infix notation $A \otimes B$ for $\otimes(A,B)$}
    \item{I, the \textbf{unit object}, is an object of $\cat{C}$}
    \item{$\alpha,\lambda,\rho$ are natural isomorphisms, with components (for all objects $X,Y,Z$ of $\cat{C}$):
    \begin{itemize}
      \item{$\alpha_{X,Y,Z}: ((X \otimes Y)\otimes Z) \rightarrow (X \otimes (Y\otimes Z))$ -- the \textbf{associator}}
      \item{$\rho_X: X \otimes I \rightarrow X$ -- the \textbf{right unitor}}
      \item{$\lambda_X: I \otimes X \rightarrow X$ -- the \textbf{left unitor}}
    \end{itemize}
    These natural isomorphisms must satisfy several coherence conditions often called the triangle and pentagon equations. Essentially, this ensures that any well-typed diagram consisting of $\otimes,\alpha,\lambda,\rho,$ $\alpha^{-1},\rho^{-1},\lambda^{-1}$ commutes.}
  \end{itemize}
  A monoidal category is further \textbf{symmetric} if there is a \textbf{braiding} natural isomorphism $\theta_{X\otimes Y}: X \otimes Y \rightarrow Y \otimes X$ such that $$\theta_{Y \otimes X} \circ \theta_{X \otimes Y} = \text{id}_{X \otimes Y}$$
\end{defn}

Monoidal categories are models of process theories \cite{bob_coecke_aleks_kissinger_picturing_2017}, and admit a sound and complete graphical calculus. The elements of the graphical calculus are as follows, from left to right of (\ref{basicprocesstheory}):
\begin{itemize}
  \item{A wire of type $A$ (the identity process on the object $A$), which ferries stuff of that type.}
  \item{A process $f$, drawn as an axis-aligned box, which takes stuff of type $A$ as input, and outputs stuff of type $B$.}
  \item{A vertical composite process of named processes $f$ and $g$. First $f$ `transforms' $A$-stuff to $B$-stuff, and then $g$ transforms that $B$-stuff into $C$-stuff.}
  \item{A horizontal composite process of named processes $h$ and $k$. Since we can name $h$ and $k$ separately, why not invoke those processes concurrently? This composite takes $A$-stuff and $C$-stuff, and returns $B$-stuff and $D$-stuff.}
  \item{A process $u$ that gives some output, but takes no input. We call these `states'. More precisely these are morphisms of type $u:I\morph{}A$ for some object $A$.}
  \item{A process $v$ that accepts some input, but gives no output. We call these `effects'. Similar to states, effects have type $v:A\morph{}I$. }
  \item{The braid natural isomorphism in a symmetric monoidal category is depicted as a pair of crossing wires.}
\end{itemize}

\begin{equation}\label{basicprocesstheory}
  \tikzfigscale{1}{figs/basicprocesstheory}
\end{equation}

The identity on the monoidal unit $I$ is represented as a blank diagram. In this work, we read diagrams bottom to top. In symmetric monoidal categories (SMCs), two diagrams are equal if they can be deformed into one another while keeping all boxes and lines on the page, and while ensuring wires never have a horizontal gradient: that is to say, they are equivalent up to planar isotopy respecting the processiveness of the wires. For example: 

\begin{equation}
  \tikzfigscale{1}{figs/compositeprocess}
\end{equation}

Since we take our braiding to be symmetric, we need not worry about the order in which wires cross each other. Nevertheless, most of this work can be easily extended to the case of non-symmetric braided monoidal categories.

The commonality of extant approaches to modelling semantics in categorical grammar is that wires are interpreted as `meaning spaces', and states on such wires correspond to specific points in those spaces. For instance, in the category \textbf{Set} of sets and functions, viewed as a SMC with cartesian product as monoidal product and singletons as monoidal unit, the wires are sets, and states on wires are specific elements of those sets. In the SMC \textbf{Vect} of vector spaces and linear maps with the standard tensor product as monoidal product, states on wires are vectors, which are used to encode semantics in distributional models of meaning.

\subsection{Spiders}
The final mathematical structure necessary for this work is that of ``spiders''.
Deeply studied by the categorical quantum mechanics line of research, spiders are algebraic structures internal to monoidal categories, usually as models of special commutative $\dag$-Frobenius algebras \cite{bob_coecke_aleks_kissinger_picturing_2017}.
We will not need all this structure, but for completeness we include those bits which will be useful for this work.

A \textit{magma} is an object $A$ with a morphism $A\otimes A\morph{} A$, depicted as a dot with legs:

\begin{equation}
  \tikzfigscale{1}{figs/magma}
\end{equation}
A magma is associative if 

\begin{equation}
  \tikzfigscale{1}{figs/associativity}
\end{equation}

A magma has a unit if there is a state, depicted as a one-legged spider, which satisfies:

\begin{equation}
  \tikzfigscale{1}{figs/unit}
\end{equation}

An associative, unital magma is also known as a \textit{monoid}, $(A,\bmult,\bunit)$.
As is typical with categorical methods one can dualise all the previous notions by reversing the direction of all the arrows.
For the diagrams this means reflecting them in the horizontal axis, giving the following notions of a comagma, coassociativity and a counit respectively:

\begin{equation*}
  \tikzfigscale{1}{figs/codefn}
\end{equation*}
overall giving a \textit{comonoid}.

Given two monoids $(A,\bmult, \bunit)$ and $(B,\wmult, \wunit)$, a map which allows one to transform between them is known as a \textit{monoid homomorphism} $h:A\morph{}B$, and must satisfy:
\begin{equation*}
  \tikzfigscale{1}{figs/monoidhom}
\end{equation*}
Comonoid homomorphisms are defined analogously, reflecting everything in the horizontal axis. We note that full spiders in the categorical quantum mechanics settings are objects that are simultaneously monoids and comonoids, that in addition satisfy the Frobenius equations, not depicted here.

\section{Axiomatic Updating}

In the following, we borrow terminology from the language of lenses, as studied in applied category theory.
Lenses were developed as formal constructs \cite{foster_combinators_nodate} to study synchronised update systems \cite{bancilhon_update_1981}, and our aims here are broadly aligned.
Our basic object of study is what we call an `update structure', which models updating an abstract system with an abstract property.
We define what qualifies as an update structure in an SMC axiomatically, reflecting three intuitive cognitive criteria for updating static properties of systems.

\begin{enumerate}
    \item Examining a property of the system and then immediately overwriting with the same property is the trivial operation.
    \item Updating an object with properties in succession is procedurally equivalent to manipulating the collection of properties in some way beforehand to obtain a single property, and then overwriting once.
    \item Repeatedly examining a property produces the same outcomes.
\end{enumerate}

\begin{defn}[Update Structure]
An update structure $(\putt{},\get{},\mix{},\copyy{})$ in a SMC consists of:
\begin{itemize}
    \item An object $S$, which we refer to as a \textbf{system}
    \item An object $p$, which we refer to as a \textbf{property}, which has:
    \begin{itemize}
    \item A magma structure $\mix{}: p \otimes p \rightarrow p$
    \item A cocommutative, coassociative comagma structure $\copyy{}: p \rightarrow p \otimes p$
    \end{itemize}
    \item A \textbf{Put} operation $\putt{}: S \otimes p \rightarrow S$
    \item A \textbf{Get} operation $\get{}: S \rightarrow S \otimes p$
\end{itemize}
Which satisfy the following equations:
\begin{itemize}
    \item (PutGet) To capture the intention of overwriting: 
    \begin{equation}\label{putget}
    \tikzfigscale{1}{figs/putget}
    \end{equation}
    \item (GetPut) To capture our first criterion, treating the identity process as trivial: 
    \begin{equation}\label{getput}
    \tikzfigscale{1}{figs/getput}
    \end{equation}
    \item (PutPut) To capture our second criterion, using the $\mix{}$ magma to encode property pre-processing: 
    \begin{equation}\label{putput}
    \tikzfigscale{1}{figs/putput}
    \end{equation}
    \item (GetGet) To capture our third criterion, following from the cocommutativity of $\copyy{}$:
    \begin{equation}\label{getget}
    \tikzfigscale{1}{figs/getget}
    \end{equation}
\end{itemize}
\end{defn}

Mathematically speaking, the PutPut axiom (\ref{putput}) says that $\putt{}$ is a magma module for $\mix{}$ while the GetGet axiom (\ref{getget}) says that $\get{}$ is a comagma comodule for $\copyy{}$.
These are weaker notions than are typically meant by a (co)module, which would usually be defined as an action over a monoid and thus also have to satisfy a unit law. 

This weakening is necessary to capture updating in some categories.
Modules and comodules in their usual setting have seen application in the categorical quantum mechanics line of research where they can describe observables and measurements \cite{coecke_measurements} and even the evolution of quantum systems under the Schr\"odinger equation \cite{gogioso_schrodinger, gogioso_monadic, gogioso_dynamics, heunen_monads}.
For further discussion of these issues, the reader is referred to \cite{wilson2020safari} where we show that quantum measurements and very-well-behaved (vwb) lenses can all be encoded as update structures.
In particular, the magma for vwb lenses is not a monoid (it does not have a two-sided unit) and thus the weakening to magma modules is necessary and interesting.

The first criterion ``examining a property of the system and then immediately overwriting with the same property is the trivial operation,'' and thus also the GetPut axiom, carries a subtle assumption: that the system is guaranteed to have the type of property read out, prior to read-out. In anticipation of this we also define a slight weakening of an update structure, one where the GetPut axiom does not hold, but where it is replaced with the \textit{repeat-update} axiom.

\begin{defn}[Weak Update Structure]
  A weak update structure $(\putt{},\get{},\mix{},\copyy{})$ satisfies all the axioms of an update structure apart from GetPut (\ref{getput}) which we replace with the strictly weaker \textit{repeat-update} axiom:
  \begin{equation}\label{compareaxiom}
    \tikzfigscale{1}{figs/compareaxiom}
  \end{equation}
\end{defn}
A weak update structure is one for which updating a system with a property it has already been updated with, does nothing. In particular this weaker axiom allows for the possibility that the system did not originally have that type of property. In section \ref{sec:weaktostrong} we will show how to ``upgrade'' any weak update structure to an update structure, in doing so we will have demonstrated that a weak update structure is a tool for building concepts.
We are justified in calling such an update ``weak'' by the following proposition:

\begin{prop}\label{prop:strongisweak}
  Any update structure is also a weak update structure.
\end{prop}
\begin{proof}
  Follows by application of axiom \ref{putget} followed by axiom \ref{getput} to the right hand side of \ref{compareaxiom}, given explicitly in the appendix.
\end{proof}

Given the semantic modeller's choice of SMC, we take the objects of that SMC to be conceptual spaces.
For instance, in the SMC of convex spaces and convexity preserving maps between them, there is an object corresponding to RGB colour-space, the states of which are particular colours.
Conceptual spaces admit concepts as subregions, as `redness' would be defined as a subregion of RGB colour space.
Generally, such concepts may be complex subregions of conceptual spaces which do not have an analogue object in the original SMC.
Moreover, concepts like `monkeyness' might be modelled in terms of other, hierarchical concepts, such as `emotion' and `banananess', where the latter is modelled in terms of primitive conceptual spaces such as `colour' and `taste'.

We may define concepts from conceptual spaces through update structures, by way of what we call `semantic correlators'.
Considering concepts as semantic correlators along with update structures provides the necessary conceptual tools to construct complex hierarchical concepts such as `monkeyness' from other concepts.

\section{Semantic Correlators}\label{sec:corup}
Given update structures for multiple properties, say, an update structure for colour and an update structure for taste, it is natural to ask how one can combine these together to create new update structures.
In particular, one might wish to build a new update structure in the presence of additional information about how two properties combine together.
For instance, if one were considering bananas then it is typical that yellow bananas are sweet and green bananas are bitter and the properties of colour and taste are no longer fully independent.
To capture such instances we introduce a new map we call the correlator.

\begin{defn}[Correlator]
  Given a family of properties $A_1, \dots, A_n, n\in\mathbb{N}$, a correlator is an idempotent map $\cor:A_1\otimes\dots\otimes A_n \morph{} A_1\otimes\dots\otimes A_n$, which is a comagma homomorphism for the comagma induced by those on each $A_i$,
  \begin{equation}
     \tikzfigscale{1}{figs/comonhom}
  \end{equation}
  and a magma homomorphism for the magma induced by those on each $A_i$
  \begin{equation}
    \tikzfigscale{1}{figs/monhom}
  \end{equation}
\end{defn}

One should think of the correlator as containing additional information about how certain properties influence each other. 
As such, correlators are very much dependent on the new concept one wishes to create from the underlying ones; the correlator for a banana is clearly very different from that of a fire engine, for instance.

The idempotence of a correlator captures the notion that properties correlated in a particular fashion are invariant under correlating them the same way again.
Demanding that a correlator be a comagma and magma homomorphism captures the notion that copying (mixing) then correlating ought to be the same as correlating then copying (mixing).

For simplicity we will now work with correlators on only two properties, but the following techniques can easily be extended to more complicated cases.
Suppose we have two properties, $c$ and $t$ with corresponding update structures $(\putt{c}, \get{c}, \mix{c}, \copyy{c})$ and $(\putt{t}, \get{t}, \mix{t}, \copyy{t})$ on a system $S$ and suppose that the structures commute, that is $\putt{t}\circ\putt{c}=\putt{c}\circ\putt{t}$ and $\get{t}\circ\get{c} = \get{c}\circ\get{t}$ where the composition is on the conceptual space $S$ and equality holds up to a braid on the properties\footnote{This notion of the \emph{independence} of the properties $c$ and $t$ reflects G\"{a}rdenfors' notion of independent domains, and is not unreasonable to require from a modelling perspective.}.
Then one can construct a new correlated Put $\putt{ct}$, Get $\get{ct}$, magma $\mix{ct}$, and comagma $\copyy{ct}$ by
%\begin{equation}
%  \tikzfigscale{1}{figs/correlatedput}, 
%  \hspace{1cm}
%  \tikzfigscale{1}{figs/correlatedget}
% \end{equation}
%\begin{equation}
%  \tikzfigscale{1}{figs/mon}
%  \hspace{1cm}
%  \tikzfigscale{1}{figs/comon}
%\end{equation}
\begin{equation}\label{correlatedupdate}
  \tikzfigscale{1}{figs/correlatedupdate}
\end{equation}
\begin{prop}\label{prop:corupdate}
 Let $\, (\putt{c}, \get{c}, \mix{c}, \copyy{c})$ and $(\putt{t}, \get{t}, \mix{t}, \copyy{t})$ be commuting update structures and $\cor$ be a correlator on $c$ and $t$.
  Then $(\putt{ct}, \get{ct}, \mix{ct}, \copyy{ct})$ defined as in (\ref{correlatedupdate}) forms a weak update structure.
\end{prop}
The GetPut axiom fails because there is no guarantee that prior to update the system has its colour and taste properties consistent with the correlator. We will be able to resolve this issue in section \ref{sec:weaktostrong} after moving to the category of semantic concepts in section \ref{sec:fruit}.
%\begin{equation*}
%  \tikzfigscale{1}{figs/correlatedputgetmaintext}
%\end{equation*}

For clarification of how the process of constructing correlators and correlated updates works consider the aforementioned example regarding the colour and taste of bananas.
The setting for our example will be $\cat{Rel}$, the category of sets and relations.
Any relation which is a function is written with function notation, any relation $R$ which cannot be written as a function is written in the form 
\begin{equation*}
  R = \{ (x,y) \ | \ xRy \}
\end{equation*}

Suppose we identify the possible colours with the set $\llbracket\textrm{colour}\rrbracket:= \{ \textrm{yellow},\ \textrm{green}\}$ and possible tastes with the set $\llbracket \textrm{taste}\rrbracket := \{\textrm{bitter}, \ \textrm{sweet}\}$ and assume for simplicity that yellow bananas are always sweet and green bananas always bitter.
Thus we have a map $P:\llbracket\textrm{colour}\rrbracket \morph{} \llbracket \textrm{taste}\rrbracket :: \textrm{yellow}\mapsto \textrm{sweet},\  \textrm{green}\mapsto\textrm{bitter}$.
Take the comultiplication on colour to be the copying map of the elements of the set of colours, $\wcomult:i\mapsto (i,i)$ for $i\in\llbracket \textrm{colour}\rrbracket$, and similarly for taste. One can then define a correlator in terms of $P$, $\wcomult$ and the dagger of the copy map, the relation $\wmult = \{((i,i),i) \ | \ i \in \llbracket\textrm{colour}\rrbracket \}$\

\begin{equation}\label{bananacor}
  \tikzfigscale{1}{figs/bananacor}
\end{equation}
which is clearly idempotent and it is straightforward to check that it is a comonoid homomorphism for the comonoid induced by copying colour and taste.
Abbreviating the properties to their first letters, the correlator can written as the following relation
\begin{equation*}
  \cor{} = \{((\textrm{y},\textrm{s}),(\textrm{y},\textrm{s})),((\textrm{g}, \textrm{b}),(\textrm{g}, \textrm{b}))\}
\end{equation*}
which throws away anything yellow-bitter or green-sweet.
Now take the following Gets, Puts and Mixes:

\begin{align*}
  \putt{c} &:: \langle c, t\rangle \times c' \mapsto \langle c' ,t \rangle & \putt{t} & :: \langle c, t\rangle \times t' \mapsto \langle c ,t' \rangle  \\
  \get{c} &:: \langle c, t\rangle \mapsto \langle c, t\rangle \times c & \get{t} & :: \langle c, t\rangle \mapsto \langle c, t\rangle \times t \\
  \mix{c} &:: c\times c' \mapsto c' & \mix{t} & :: t\times t' \mapsto t'
\end{align*}
for $c,c'\in \llbracket\textrm{colour}\rrbracket$ and $t,t'\in \llbracket\textrm{taste}\rrbracket$.

Then one can confirm that $\cor$ (\ref{bananacor}) is a monoid homomorphism for the monoid induced by $\mix{c}$ and $\mix{t}$.
The correlated weak update for which green bananas are bitter and yellow bananas sweet is constructed by composing the above relations.

\section{Fruit Wires: The Category of Semantic Concepts}\label{sec:fruit}
Taking the view that concepts can be defined by the correlations they exhibit between properties in independent domains, we aim to construct the concept $banana$ (\banana) from the domains $colour$ ($c$) and $taste$ ($t$) along with the correlator \[\cor_{\smallbanana}: c \otimes t \rightarrow c \otimes t\]
Informally, any state $\psi$ on the property pair $c \otimes t$ which is consistent with the correlator should represent an instance of a banana. The formal condition below is that banana-states $\psi$ are those that satisfy:
\begin{equation}
  \tikzfigscale{1}{figs/karoubis}
\end{equation}
Additionally the processes $f$ that can be performed on bananas should be all of those which are consistent with the correlations: applying the correlator to the inputs and outputs of $f$ should have no effect. Formally this means we consider every admissible banana-process $f$ to satisfy:
\begin{equation}
  \tikzfigscale{1}{figs/karoubif}
\end{equation}
%The process of learning about a banana then consists of first identifying the basic properties of a banana which in our toy example we take to be color and taste $C,T$ and then restricting the processes on $C \otimes T$ to be only those consistent with the correlator $Cor$. 

One can capture such scenarios by transitioning from working in the category $\cat{C}$ to the Karoubi envelope $\bar{\cat{C}}$.
\begin{defn}[Karoubi Envelope]
  The Karoubi envelope $\bar{\cat{C}}$ of a category $\cat{C}$ has as objects the pairs $(A,\pi)$ where $A \in \ob{\cat{C}}$ and $\pi: A \rightarrow A$ is an idempotent.
  The morphisms $f : (A ,\pi) \rightarrow (B,\sigma)$ are the morphisms $f:A \rightarrow B$ such that $\sigma \circ f = f = f \circ \pi$. 
\end{defn}
Objects of the form $(c \otimes t, \cor_{\smallbanana})$ in the Karoubi envelope $\bar{\cat{C}}$ are precisely those for which the only allowed processes are those consistent with $\cor_{\smallbanana}$ and thus it makes sense to define new concepts as objects of this form for a given correlator.

\begin{defn}[Concept]
The concept corresponding to the correlator $\cor: (A,\pi_{A}) \otimes (B,\pi_{B}) \rightarrow (A,\pi_{A}) \otimes (B,\pi_{B})$ is the object $(A \otimes B, \cor)$ in $\bar{\cat{C}}$.
\end{defn}
We interpret the Karoubi envelope $\bar{\cat{C}}$ to be the category of semantic concepts, built from a modelling category of basic concepts $\cat{C}$.
It is straightforward to show that there is an equivalence of categories between the full subcategory of $\bar{\cat{C}}$ spanned by objects of the form $(A,\id{A})$ for all $A\in\ob{\cat{C}}$ and the original category $\cat{C}$.
Thus all of our initial conceptual spaces appear in $\bar{\cat{C}}$ alongside many new derived concepts given by the action of correlators.
This allows us treat new concepts as formally different types within our category.
Such notions have also been explored in the categorical quantum mechanics line of research where idempotent decoherence maps can be used to give quantum and classical data different types \cite{coecke_classicality, selinger_idempotents, Heunen_cp}.

An example of a non-basic concept is the $banana$ object $\banana = (c \otimes t, \cor_{\smallbanana})$, built from a correlator \[\cor_{\smallbanana}: (c,\id{c}) \otimes (t,\id{t}) \rightarrow (c,\id{c}) \otimes (t,\id{t})\] upon the (basic) conceptual spaces $colour$ $(c,\id{c})$ and $taste$ $(t,\id{t})$.
Intuitively $banana$ wires are $colour$ and $taste$ wires which have been correlated so as to be consistent with the concept $banana$.
\begin{equation}
  \tikzfigscale{1}{figs/bananawire}
\end{equation}
This construction iterates, allowing definition of derived concepts which admit other derived concepts as properties.
A further correlation between the flavour (and thus colour) of bananas and the happiness of a consumer induces its own correlator:
\begin{equation}\label{monkeycor}
  \cor_{\smallmonkey} : (\banana,\cor_{\smallbanana}) \otimes (\smiley,\cor_{\smallsmiley}) \rightarrow (\banana,\cor_{\smallbanana}) \otimes (\smiley,\cor_{\smallsmiley})
\end{equation}
which acts on the properties \textit{banana} and \textit{emotion}.
Such a correlator could be used to construct the concept \textit{monkey} $\monkey \equiv (\banana \otimes \smiley, \cor_{\smallmonkey})$ as the collection of maps which are consistent with the new correlator (\ref{monkeycor}).
The consistency of states and processes on $\monkey$ with its constituent properties $\banana$ and $\smiley$ is guaranteed since for any $f: \monkey \rightarrow \monkey$, \[ f \circ (\cor_{\smallbanana} \otimes \cor_{\smallsmiley}) = f \circ \cor_{\smallmonkey} \circ (\cor_{\smallbanana} \otimes \cor_{\smallsmiley}) \] \[ =  f \circ \cor_{\smallmonkey} =  \cor_{\smallmonkey} \circ f  = (\cor_{\smallbanana} \otimes \cor_{\smallsmiley}) \circ \cor_{\smallmonkey} \circ f  = (\cor_{\smallbanana} \otimes \cor_{\smallsmiley}) \circ f\]

In the previous section we demonstrated that a correlator can be used to construct a (weak) correlated update in $\cat{C}$ and we have now demonstrated that a correlator in $\cat{C}$ can be used to construct a new concept in $\bar{\cat{C}}$. These two constructions can be brought together by noting that the morphisms in $\cat{C}$ on $S,c$ and $t$ of a correlated update $(\putt{ct}, \get{ct}, \mix{ct}, \copyy{ct})$ are indeed morphisms in $\bar{\cat{C}}$ on $(S,\id{S})$ and $(c \otimes t, \cor{})$. As such $(\putt{ct}, \get{ct}, \mix{ct}, \copyy{ct})$ defines a weak update structure in $\bar{\cat{C}}$ on system $(S,\id{S})$ and correlated property $(c\otimes t, \cor{})$.

To summarise, we may then treat $\banana$ as a property to be read out from system $(S,\id{S})$ by replacing updates for the basic properties $colour$, and $taste$ with an update for the property $banana$ $\banana = (c \otimes t, \cor_{\smallbanana})$ on $(S,\id{S})$;
\begin{equation}
  \tikzfigscale{1}{figs/karoubiput}
\end{equation}
Whilst the values read-out by a correlated update structure will be consistent with $\cor{}$ there is no guarantee that the system itself actually has its colour and taste properties consistent with the property $\cor{}$, for the latter we will need a system on which the GetPut axiom holds.
%Now that we have developed a way of constructing new concepts from old ones we are able to return to the problem of combining update structures and state the following proposition:

%\begin{prop}\label{prop:corupdate}
%  Let $\, (\putt{c}, \get{c}, \mix{c}, \copyy{c})$ and $(\putt{t}, \get{t}, \mix{t}, \copyy{t})$ be commuting update structures and $\cor$ be a correlator on $c$ and $t$.
%  Then $(\putt{ct}, \get{ct}, \mix{ct}, \copyy{ct})$ forms a weak update structure on correlated inputs $(c\otimes t, \cor{})$ where $\putt{ct}$ and $\get{ct}$ are defined as in (\ref{correlatedupdate}) and $\mix{ct}$ and $\copyy{ct}$ are the magma and comagma induced by those on $c$ and $t$.
%\end{prop}

\section{From Weak to Strong Update Structures}\label{sec:weaktostrong}
In the previous section, we demonstrated how to construct a correlated colour and taste concept $(c \otimes t, \cor{})$, along with the corresponding weak update structure on $(S,\id{S})$ and $(c \otimes t , \cor{})$.
Here we show how to use this weak update structure to construct a new system concept $(S,\pi_S')$ on which all states have correlated colour and taste properties. 
%Put formally: a weak update structure can be used to enforce that an arbitrary concept, viewed as a system $(S,\pi_S)$, has properties ${c}$ and ${t}$ consistent with a (derived) property $(b,\pi_b) = (c \otimes t, \cor)$, by constructing from a weak update $(\putt{b}, \get{b}, \mix{b}, \copyy{b})$ of the property $(b,\pi_b)$ on the system $(S,\pi_S)$ a new system $(S,\pi_S ')$, along with a strong update structure on that system. 
 
%To begin we examine the \textit{repeat-update} axiom, which is a weakening of the GetPut axiom. They are, respectively:
%The \textit{repeat-update} axiom, satisfied by the correlated update defined in (\ref{correlatedupdate}), is a statement of static-ness of the update property.
%Updating with `copies' of a property ${b}$ produced by the comagma is the same as updating with that property once.
%The GetPut axiom $\putt{ct} \circ \get{ct} = \id{(S,\id{})}$ is (in the presence of axioms \ref{putget}, \ref{putput}, \ref{getget}) strictly stronger than the \textit{repeat-update} axiom; intuitively it is the additional claim that the system $(S,\pi_S)$ actually has colour and taste properties consistent with $(c \otimes t, \cor{})$. For example 
The failure of GetPut for a correlated update as defined in (\ref{correlatedupdate}) comes about because the system was originally assumed to have independent colour and taste values, some of which would be inconsistent with (and so affected by) the correlator.
The GetPut morphism $\putt{ct} \circ \get{ct}$ represents the process of extraction, correlation, and finally re-insertion, of colour and taste properties.
Since the process of correlation can have a non trivial effect on the system the GetPut morphism $\putt{ct} \circ \get{ct}$ will not be the identity. 
\begin{equation}
  \tikzfigscale{1}{figs/failstronggetput2}
\end{equation}
An informal way to enforce that properties ${c}$ and ${t}$ of $(S, \id{S})$ are correlated is to first apply the GetPut morphism $\putt{ct} \circ \get{ct}$ to the system $(S,\id{S})$, thereby correlating its properties.
Formally, an implication of the \textit{repeat-update} axiom is that $\putt{ct} \circ \get{ct}$ is itself an idempotent,
\begin{equation*}
  \putt{{ct}} \circ \get{{ct}} \circ \putt{{ct}} \circ \get{{ct}} = \putt{{ct}} \circ \get{{ct}}
\end{equation*}
%and thus applying $\putt{ct} \circ \get{ct}$ to a system which is already consistent with the correlations $\cor$ will have no effect.
%\begin{equation}
%\tikzfigscale{1}{figs/getputidempotent}
%\end{equation}
and so there exists an object in $\bar{\cat{C}}$ which enforces consistency of the system with $\cor$ \begin{equation*}
  (S,\pi_{S}') \equiv (S,\putt{ct} \circ \get{ct})
\end{equation*}
Any state $\psi: (I,\id{I}) \rightarrow (S,\putt{ct} \circ \get{ct})$ already has correlated colour and taste properties in  the sense that by definition it satisfies $\putt{ct} \circ \get{ct} \circ \ \psi = \psi$.
What remains is to construct a new update structure on system $(S,\pi_{S}')$ and property $(c \otimes t, \cor)$, satisfying the GetPut axiom $\putt{{ct}} \circ \get{ct} = \id{(S,\pi_{S}')}$.
In the following proposition we allow the original system to have the form $(S,\pi_S)$ as opposed to the restricted form $(S,\id{S})$, and we allow the property to have the general form $(b, \pi_b)$ as opposed to the restricted form $(c \otimes t, \cor{})$.

\begin{prop}[GetPut Restriction]\label{reptoget}
  For a weak update structure $(\putt{b}, \get{b}, \mix{b}, \copyy{b})$ on system $(S,\pi_{S})$ and property $(b,\pi_b)$ satisfying the \textit{repeat-update} axiom, the GetPut-Restricted updates
  \begin{equation}
    \tikzfigscale{1}{figs/newgetandput3}
  \end{equation}
  along with $(\mix{b}, \copyy{b})$ define an update structure on system $(S,\putt{b} \circ \get{b})$ and property $(b,\pi_b)$. 
\end{prop}
\begin{proof}
  By idempotency of $\putt{b} \circ \get{b}$ it is immediate that $\get{b}'$ and $\putt{b}'$ appear as morphisms 
  \begin{equation*}
    \get{b}' : (S,\putt{b} \circ \get{b}) \rightarrow (S,\putt{b} \circ \get{b}) \otimes (b,\pi_b)
  \end{equation*} 
  and 
  \begin{equation*}
    \putt{{b}}' :  (S,\putt{b} \circ \get{b}) \otimes (b,\pi_b) \rightarrow (S,\putt{b} \circ \get{b})
  \end{equation*} respectively.
  The update structure axioms \ref{putget}, \ref{getput}, \ref{putput}, \ref{getget} all hold, the proofs of which are given in the appendix.
\end{proof}
The statement and proof of proposition \ref{reptoget} makes no explicit reference to correlations, and so gives a general way to force consistency of a system with its properties by moving from a weak update structure satisfying the \textit{repeat-update} axiom, to an update structure satisfying the GetPut axiom. 
%This idea has applications to update structures in the quantum mechanical setting, when classical agents measure quantum systems, they ask for properties that systems do not a-priori have \cite{}.

For clarification of the above process we build on the basic explicit example in $\cat{Rel}$ given in section \ref{sec:corup}.
We take a system set $S$ of the form $\llbracket \textrm{emotion}\rrbracket \times \llbracket \textrm{colour}\rrbracket \times \llbracket \textrm{taste}\rrbracket$, being used to model a monkey.
We can immediately define an update structure for each of colour and taste. 
\begin{align*}
  \putt{c} &:: \langle e, c, t\rangle \times c' \mapsto \langle e,   c' ,t \rangle & \putt{t} & :: \langle e, c, t\rangle \times t' \mapsto \langle e, c ,t' \rangle  \\
  \get{c} &:: \langle e, c, t\rangle \mapsto \langle e, c, t\rangle \times c & \get{t} & :: \langle e, c, t\rangle \mapsto \langle e, c, t\rangle \times t \\
  \mix{c} &:: c\times c' \mapsto c' & \mix{t} & :: t\times t' \mapsto t'
\end{align*}
with $c,c'\in \llbracket\textrm{colour}\rrbracket$ and $t,t'\in \llbracket\textrm{taste}\rrbracket$.

The GetPut relation
\begin{equation*}
  \putt{ct} \circ \get{ct}: \llbracket \textrm{emotion}\rrbracket \times \llbracket \textrm{colour}\rrbracket \times \llbracket \textrm{taste}\rrbracket \rightarrow \llbracket \textrm{emotion}\rrbracket \times \llbracket \textrm{colour}\rrbracket \times \llbracket \textrm{taste}\rrbracket
\end{equation*}
can now be explicitly computed, it acts point-wise as \begin{equation*}
  \putt{ct} \circ \get{ct} = \bigcup_{e \in \llbracket \textrm{emotion}\rrbracket} \{((\textrm{e},\textrm{y},\textrm{s}),(\textrm{e},\textrm{y},\textrm{s})),((\textrm{e},\textrm{g}, \textrm{b}),(\textrm{e}, \textrm{g}, \textrm{b}))\}
\end{equation*}
Any state $\psi: (I,\id{I}) \rightarrow (S, \putt{ct} \circ \get{ct})$ must then be a set of elements of $S$ unaffected by the throwing away of yellow-bitter and green-sweet components, i.e.
\begin{equation*}
  \psi \subseteq \bigcup_{e \in \llbracket \textrm{emotion}\rrbracket} \{(\textrm{e},\textrm{y},\textrm{s}),(\textrm{e},\textrm{g},\textrm{b})\}
\end{equation*}
By the form of the states $\psi: (I,\id{I}) \rightarrow (S, \putt{ct} \circ \get{ct})$ we can see that the system $(S,\putt{ct} \circ \get{ct})$ indeed represents a system for which the colour and taste properties are guaranteed to be consistent with the correlator.

\section{Conclusion}

We have demonstrated a proof-of-concept for the categorical representation of concepts as restricted and correlated regions of G\"{a}rdenfors' conceptual spaces, which, to our knowledge, is the first semantic-engineering contribution of its kind: a cognitively driven axiomatics for a formal representational and computational model of mental concepts.

The core element of our construction is an interacting magma module and comagma comodule, internal to a symmetric monoidal category, reflecting `update' and `examine' properties.
Though we borrowed terminology from lenses, our approach is distinct on two counts.
Firstly, we define our constructs in arbitrary symmetric monoidal categories, whereas lenses typically assume cartesian categories.
Secondly, the notion of interacting (co)magma modules is conceptually minimal, there being no analogue in the field of profunctor optics at large to the update structures we have defined, to the best of our knowledge.
The benefit of this conceptual minimalism is that the requirements for update structures to exist are readily met by most approaches to categorical linguistic models. In this work, we have only considered $\cat{Rel}$-based semantic models, but there is no obstacle in principle to applying the same approach in true convex geometric domains and to vector-representation semantics.

It remains for future work to demonstrate practical applications, say, in terms of inference in vector-based models of linguistic meaning, where we expect our update formalism will permit dynamic epistemic modelling as is encountered in the reading of a long text \cite{coecke_mathematics_2019}, where the properties of various actors change over time.
It is also pressing work to integrate our approach with extant research on density-matrix based models of compositional cognition \cite{al-mehairi_compositional_2016}.

A further added benefit of such a conceptually minimal and axiomatic approach is that new domains of semantic models are now amenable for use, for instance, those obtained by RDF triples: no matter the underlying representations, once one has verified that a candidate update structure satisfies the axioms, one can be confident that the structure suitably models updating and viewing properties.

We are surprised by the power and fruitfulness of adopting graphical axiomatics that reflect cognitive modelling intentions, and we believe that further progress can be made in semantic engineering following this general method.
We have begun to explore various other axiomatisations of update structures \cite{wilson2020safari}, each capturing different operational intuitions, and yielding qualitatively different systems, but it remains for future work to fully tame the zoo of update structures obtained in this way.

\subsubsection*{Acknowledgements}
With thanks to Sean Tull, Cole Comfort, Irene Rizzo, and Bob Coecke for helpful discussions.
This work was supported by University College London and the EPSRC
[grant number EP/L015242/1].
The authors contributed equally to this work.

\bibliographystyle{eptcs}
\bibliography{putget}

\appendix
\section{Proofs}
\subsection{Proof of Proposition \ref{prop:strongisweak}}
\begin{proof}
By application of PutGet followed by GetPut
\begin{equation*}
\tikzfigscale{1}{figs/strongtoweak}
\end{equation*}
\end{proof}
\subsection{Proof of Proposition \ref{prop:corupdate}}
\begin{proof}
We check the PutPut, GetGet, PutGet and \textit{repeat-update} axioms, recalling that we assume that the Gets and Puts of the two systems respectively commute. PutPut follows by:
\begin{equation*}
\tikzfigscale{1}{figs/correlatedputput}
\end{equation*}
\begin{equation*}
\tikzfigscale{1}{figs/correlatedputput2}
\end{equation*} up to correlating the
with GetGet very similar, PutGet follows:
\begin{equation*}
\tikzfigscale{1}{figs/correlatedputget}
\end{equation*}
\begin{equation*}
\tikzfigscale{1}{figs/correlatedputget2}
\end{equation*}
Finally, GetPut, does not hold on all input systems:
\begin{equation*}
\tikzfigscale{1}{figs/correlatedgetput}
\end{equation*}
but the \textit{repeat-update} axiom is easy to check.
\end{proof}

\subsection{Proof of Proposition \ref{reptoget}}
Making heavy use of the equation,
\begin{equation*}
\tikzfigscale{1}{figs/usefulidentity}
\end{equation*}
along with idempotency of $\putt{{b}}' \circ \get{{b}}'$ throughout, we begin by verifying PutPut,
\begin{equation*}
\tikzfigscale{1}{figs/proveputput}
\end{equation*}
GetPut
\begin{equation*}
\tikzfigscale{1}{figs/PROVEGETPUT}
\end{equation*}
PutGet
\begin{equation*}
\tikzfigscale{1}{figs/proveputget}
\end{equation*}

Finally, since
\begin{equation*}
\tikzfigscale{1}{figs/proveggpart1}
\end{equation*}
\begin{equation*}
\tikzfigscale{1}{figs/proveggpart2}
\end{equation*}
GetGet follows
\begin{equation*}
\tikzfigscale{1}{figs/provegetgetpart2}
\end{equation*}

\end{document}